\documentclass[conference]{IEEEtran}
\pdfoutput=1
\usepackage{times}

\usepackage[numbers]{natbib}
\usepackage{multicol}
\usepackage{amsmath}
\usepackage{amsfonts}
\usepackage{amssymb}
\usepackage{graphicx}
\usepackage{listings}
\usepackage{colortbl}

\definecolor{darkred}{RGB}{200,0,0}
\definecolor{darkgreen}{RGB}{0,127,0}
\definecolor{purple}{RGB}{127,0,127}
\definecolor{shadow}{RGB}{150,150,150}

\newcommand{\ex}{\mathbb{E}}

\newcommand{\x}{\mathbf{x}}

\newcommand{\I}{\mathbf{I}}
\newcommand{\K}{\mathbf{K}}

\newcommand{\W}{\mathbf{W}}

\newcommand{\w}{\mathbf{w}}

\newcommand{\NP}{\mathcal{NP}}
\newcommand{\GP}{\mathcal{GP}}
\newcommand{\TP}{\mathcal{TP}}
\newcommand{\IG}{\mathcal{IG}}
\newcommand{\N}{\mathcal{N}}

\begin{document}

\title{BayesOpt: A Library for Bayesian optimization with Robotics Applications}

\author{\authorblockN{Ruben Martinez-Cantin} 
       \authorblockA{Centro Universitario de la Defensa, Zaragoza\\
         Email: rmcantin@unizar.es \\
         Zaragoza, 50090, Spain}}


%

\maketitle

\begin{abstract}
The purpose of this paper is twofold. On one side, we present a general framework for Bayesian optimization and we compare it with some related fields in active learning and Bayesian numerical analysis. On the other hand, Bayesian optimization and related problems (bandits, sequential experimental design) are highly dependent on the surrogate model that is selected. However, there is no clear standard in the literature. Thus, we present a fast and flexible toolbox that allows to test and combine different models and criteria with little effort. It includes most of the state-of-the-art contributions, algorithms and models. Its speed also removes part of the stigma that Bayesian optimization methods are only good for ``expensive functions''. The software is free and it can be used in many operating systems and computer languages. 
\end{abstract}

\IEEEpeerreviewmaketitle

\section{Introduction}
Many problems in engineering, computer science, robotics., require to find the extremum of a real valued function. In many cases, those functions do not have a closed-form expression or might be multimodal, where some of the local extrema might have a bad outcome compared to the global extremum, or the evaluation of those functions might be costly.

Global optimization is a special case of non-convex optimization where we want to find the global extremum of a real valued function, that is, the target function. The search is done by some pointwise evaluation of the target function.

The objective of a global optimization algorithm is to find the sequence of points
\begin{equation}
x_n \in \mathcal{A} \subset \mathbb{R}^m , \;\;\; n = 1,2,\ldots
\end{equation}
which converges to the point $x^*$, that is, the extremum of the target function, when $n$ is large. The algorithm should be able to find that sequence at least for all functions from a given family. 

As explained in \cite{Mockus94}, this search procedure is a sequential decision making problem where point at step $n+1$ is based on decision $d_n$ which considers all previous data:
\begin{equation}
x_{n+1} = d_n(x_{1:n},y_{1:n})
\end{equation}
where $y_i = f(x_i) + \epsilon_i$. For simplicity, many works assume $\epsilon_i = 0$, that is, function evaluations are deterministic. However, we can easily extend the description to include stochastic functions (e.g.: homoscedastic noise $\epsilon_i \sim \N(0,\sigma)$).

The search method is the sequence of decisions $d = {d_0,\ldots, d_{n-1}}$, which leads to the final decision $x_{n} = x_{n}(d)$. In most applications, the objective is to optimize the response of the final decisions. Then, the criteria relies on the \emph{optimality error} or \emph{optimality gap}, which can be expressed as:
\begin{equation}
\delta_n(f,d) = f\left(x_n\right) - f(x^*) \label{eq:opt-error}
\end{equation}
In other applications, the objective may require to converge to $x^*$ in the input space. Then, we can use for example the \emph{Euclidean distance error}:
\begin{equation}
\delta_n(f,d) = \|x_n - x^*\|_2 \label{eq:dist-error}
\end{equation}
Equations \eqref{eq:opt-error} and \eqref{eq:dist-error} can also be interpreted as variants of the \emph{loss} function for the decision at each step. Thus, the optimal decision is defined as the function that minimizes the loss function:
\begin{equation}
  \label{eq:decision}
d_n = \arg \min_d \delta_n(f,d)
\end{equation}
This requires full knowledge of function $f$, which is unavailable. Instead, let assume that the target function $f = f(x)$ belongs to a family of functions $f \in F$, e.g.: continuous functions in $\mathbb{R}^m$. Let also assume that the function can be represented as sample from a probability distribution over functions $f \sim P(f)$. Then, the best response case analysis for the search process is defined as the decision that optimizes the expectation of the loss function:
\begin{equation}
d^{BR}_n = \arg \min_d \ex_{P(f)} \left[ \delta_n(f,d)\right]= \arg \min_d \int_F \delta_n(f,d) \; dP(f) \label{eq:average}
\end{equation}
where $P$ is a prior distribution over functions. 

However, we can improve equation \eqref{eq:average} considering that, at decision $d_n$ we have already \emph{observed} the actual response of the function at $n-1$ points, $\{x_{1:n-1},y_{1:n-1}\}$. Thus, the prior information of the function can be updated with the observations and the Bayes rule:
\begin{equation}
  \label{eq:bayes}
  P(f|x_{1:n-1},y_{1:n-1}) = \frac{P(x_{1:n-1},y_{1:n-1}|f) P(f)}{P(x_{1:n-1},y_{1:n-1})}
\end{equation}
In fact, we can actually rewrite the equation to represent the updates sequentially:
\begin{equation}
  \label{eq:bayes-sec}
P(f|x_{1:i},y_{1:i}) = \frac{P(x_{i},y_{i}|f) P(f|x_{1:i-1},y_{1:i-1})}{P(x_{i},y_{i})},
\end{equation}
$\forall \; i=1 \ldots n-1$. Thus, equation \eqref{eq:average} can be rewritten as:
\begin{equation}
  \begin{split}
d^{BO}_n &= \arg \min_d \ex_{P(f|x_{1:n-1},y_{1:n-1})} \left[ \delta_n(f,d)\right]\\
        &= \arg \min_d \int_F \delta_n(f,d) \; dP(f|x_{1:n-1},y_{1:n-1})     
  \end{split}
\label{eq:bayes-average}
\end{equation}
Equation \eqref{eq:bayes-average} is the root of \emph{Bayesian optimization}, where the Bayesian part comes from the fact that we are computing the expectation with respect to the posterior distribution, also called \emph{belief}, over functions. Therefore, Bayesian optimization is a memory-based optimization algorithm.

As commented before, most of the theory of Bayesian optimization is related to deterministic functions, we consider also stochastic functions, that is, we assume there might be a random error in the function output. In fact, evaluations can produce different outputs if repeated. In that case, the target function is the expected output. Furthermore, in a recent paper by \cite{Gramacy2012} it has been shown that, even for deterministic functions, it is better to assume certain error in the observation. The main reason being that, in practice, there might be some mismodelling errors which can lead to instability of the recursion if neglected.

\section{Related fields}

Equation \eqref{eq:bayes-average} is, in fact, much more general than Bayesian optimization. There are many fields of research that combines optimal decision making over the expectation of an unknown function which is recursively learned. 

\subsection{Active and interactive learning}

Active learning consider the situation where data points can be selected for labeling during the training process. The rationale behind that is that: data points might be expensive to label, \emph{bad} data points might introduce important bias, etc. The analogy with Bayesian optimization is clear.

In fact, one can rethink Bayesian optimization as an active learning problem where the unknown parameters are the extremum of the function, i.e. $\Theta=x^*$. In this setup, equation (\ref{eq:opt-error}) correspond to the prediction bias and equation (\ref{eq:dist-error}) corresponds to the parameter variance.

In fact, the first applications of Bayesian optimization in fields like robotics or computer graphics design are all related to problems of active learning like \cite{MartinezCantin07RSS} or interactive learning \cite{Brochu07NIPS}. See Section \ref{sec:robot-appl} for a detailed description.

\subsection{Experimental design}
The field of Bayesian optimization shares important pieces of knowledge with the \emph{sequential experimental design} field, especially related to the Design and Analysis of Computer Experiments (DACE), pioneered by \cite{Sacks89SS} and beautifully reviewed in \cite{Santner03}. The main difference resides on the optimum point $x^*$, which in the case of DACE, it is not longer the extremum of the function $f$, but the point that provides most information to reconstruct the target function $f$. The reader should note also the analogy with \emph{active learning}, where the target function $f$ corresponds to the regression function or classifier that the system is learning, while trying to sequentially querying the most informative point $x^*$.

For the sequential experimental design equation \eqref{eq:opt-error} is replaced by a different loss function. For example, if we consider that the function can be represented in a --potentially infinite-- parametric form $f(x)= g(\psi,x)$, then we can define the loss function in terms of the parameters:
\begin{LaTeXdescription}
\item[A-optimality (error minimization)]
  \begin{equation}
    \label{eq:aopt}
    \delta_{AO}(f,d) = (\psi(f)-\hat{\psi})^T \Sigma_\psi (\psi(f)-\hat{\psi})  
  \end{equation}
\item[D-optimality (entropy minimization)]
  \begin{equation}
    \label{eq:dopt}
    \delta_{DO}(f,d) = \mathcal{H}(\psi|f, x_n)  
  \end{equation}
\item[I-optimality (prediction error minimization)]
  \begin{equation}
    \label{eq:iopt}
    \delta_{IO}(f,d) = (f-\hat{f})^T \Sigma_f (f-\hat{f})  
  \end{equation}
\end{LaTeXdescription}
There are related criteria (E-optimality, C-optimality, \ldots) which can be easily mapped in a loss function. For a complete description of the different criteria, see \cite{Atkinson2007}. Thus, the only difference between code for experimental design and Bayesian optimization is the criteria used.

\subsection{Reinforcement learning}

The field of reinforcement learning \cite{Kaelbling96jmlr}

\subsubsection{Multi-armed bandits}
The simplest known reinforcement learning problem is called the \emph{multi-armed bandit} problem \cite{Srinivas10}. In the bandits problem, the response function is a reward or cost function --for example, the cost of a manufacturing a product in a certain way--. The algorithm must find the lowest production cost, regarding that changing the process may result in a more expensive manufacturing cost.

The main different with the optimization problem is that intermediate evaluations of the function may incur in extra cost. Thus, the definition of the performance function is no longer the optimality error. Instead, equation \eqref{eq:opt-error} is called the \emph{instantaneous regret} $r_n$  in the reinforcement learning literature. The target of the bandits algorithm is to minimize the \emph{accumulated regret} $R_N = \sum^N_{n=1} r_n$ or \emph{average regret} $R_N/N$. Thus, \eqref{eq:opt-error} can be replaced by:
\begin{equation}
  \label{eq:bandits}
\delta_{SB}(f,d) = \frac{\sum_{n=1}^N f\left(x_{n}(d)\right) -f(x^*)}{N}    
\end{equation}
In this setup, an algorithm with \emph{no-regret}, that is $\lim_{N\rightarrow\infty} R_N/N = 0$, is guaranteed to converge to the optimum. In fact, as cleverly found by \cite{Srinivas10}, the average regret $R_N/N$ can be used to provide convergence rates of the optimization algorithm.

Typically, in the multi-armed bandit setup, each input has associated a probability distribution to deliver the reward or cost values. Thus, several evaluations of the same input results in different outcomes. The analogous case would be the optimization problem of stochastic function.

Finally, Bull \cite{Bull2011} suggest $\epsilon$-greedy, a classical bandits algorithm, to improve the convergence of Bayesian optimization.

\subsubsection{Partially-observable Markov Decision Processes (POMDP)}
As pointed out in \cite{ToussaintBSG}, Bayesian optimization, as a sequential decision making problem, has a direct connection to a POMDP. In fact, bandits can be modeled as a single state reinforcement learning problem, for example, an POMDP with immediate reward.

An independent connection between Bayesian optimization and POMDPs is through the specific methodology to solve POMDPs based on \emph{policy search}. In policy search, it is assume that the policy space can be expressed as a parametric function. In an abuse of notation, let us name the policy parameters as $\pi$. Thus, the parametrization inherently encodes the dynamics of the world (priors, transitions, etc). Then, the reinforcement learning problem becomes a \emph{static} optimization problem.
\begin{equation}
  \label{eq:apl}
  \pi^* = \arg \max_\pi \ex \left( R(\pi) \right)
\end{equation}
which is equivalent to \eqref{eq:average} replacing $\max$ by $\min$ and \emph{reward function} by \emph{loss function}.

\subsection{Bayesian numerical analysis}

In the seminal works by \cite{Diaconis1988} and \cite{O'Hagan1992} introduce the field of Bayesian numerical analysis. The main idea is to solve a complex analysis problem --for example: interpolation, regression, integral evaluation, etc.-- by following a simple methodology:
\begin{enumerate}
\item Put a prior on the family of functions (e.g.: continuous) on the working domain.
\item Compute the response of the function $f$ as a set of sample points $x_1,x_2,\ldots, x_n$.
\item Compute the posterior.
\item Solve the original problem by the Bayes rule.
\end{enumerate}

\cite{Diaconis1988} presents a set of beautiful examples where, the previous \emph{algorithm} results a powerful machinery. \cite{O'Hagan1992} goes one step further and formalize the previous methodology in a single, although very general model, showing its applicability in several analysis problems --including optimization--. In fact, all the previous fields, including Bayesian optimization itself, can be seen as particular cases of some Bayesian numerical analysis.

The models presented in this paper are partially based on \cite{O'Hagan1992} formulation.

\section{Bayesian optimization general model}

In order to simplify the description, we are going to use a special case of Bayesian optimization model defined previously which corresponds to the most common application. In subsequent Sections we will introduce some generalizations for different applications.

Without loss of generality, consider the problem of finding the minimum of an unknown real valued function $f:\mathbb{X} \rightarrow \mathbb{R}$, where $\mathbb{X}$ is a compact space, $\mathbb{X} \subset \mathbb{R}^d, d \geq 1$. Let $P(f)$ be a prior distribution over functions represented as a stochastic process, for example, a Gaussian process $\xi(\cdot)$, with inputs $x \in \mathbb{X}$ and an associate kernel or covariance function $k(\cdot,\cdot)$. Let also assume that the target function is a sample of the stochastic process $f \sim \xi(\cdot)$.

In order to find the minimum, the algorithm has a maximum budget of $N$ evaluations of the target function $f$. The purpose of the algorithm is to find optimal decisions that provide a better performance at the end, according to equation \eqref{eq:bayes-average}. 

One advantage of using Gaussian processes as a prior distributions over functions is that new observations of the target function $(x_i,y_i)$ can be easily used to update the distribution over functions. Furthermore, the posterior distribution is also a Gaussian process $\xi_i = \left[ \xi(\cdot) | x_{1:i},y_{1:i} \right]$. Therefore, the posterior can be used as an informative prior for the next iteration in a recursive algorithm.

In a more general setting, many authors have suggested to modify the standard zero-mean Gaussian process for different variations that include semi-parametric models \cite{Huang06,Handcock1993,Jones:1998,O'Hagan1992}, use of hyperpriors on the parameters \cite{MartinezCantin09AR,Brochu:2010c,Hoffman2011}, Student t processes \cite{Gramacy_Polson_2009,Sacks89SS,Williams_Santner_Notz_2000}, etc.

We use a generalized linear model of the form:
\begin{equation}
  \label{eq:genmodel}
  f(x) = \phi(\x)^T \w + \epsilon(\x)
\end{equation}
where
\begin{equation}
  \label{eq:gpasnoise}
  \epsilon(\x) \sim \NP \left( 0, \sigma^2_s (\K(\theta) + \sigma^2_n \I) \right)
\end{equation}
The term $\NP$ means a non-parametric process, which can make reference to a Gaussian process $\GP$ or a Student's t process $\TP$. In both cases, $\sigma^2_n$ is the observation noise variance, sometimes called nugget, and it is problem specific. Many authors decide to fix this value $\sigma^2_n = 0$ when the function $f(x)$ is deterministic, for example, a computer simulation. However, as cleverly pointed out in \cite{Gramacy2012}, there might be more reasons to include this term appart from being the observation noise, for example, to consider model inaccuracies.

This model has been presented in different ways depending on the field where it was used:
\begin{itemize}
\item As a generalized linear model $\phi(\x)^T\w$ with heteroscedastic perturbation $\epsilon(\x)$.
\item As a nonparametric process of the form $\NP \left(\phi(\x)^T\w, \sigma^2_s (\K(\theta) + \sigma^2_n \I) \right)$.
\item As a semiparametric model $f(\x) = f_{par}(\x) + f_{nonpar}(\x) = \phi(\x)^T\w + \NP(\cdot)$
\end{itemize}


\section{Computing the hyperparameters}
\label{sec:comp-hyperp}

Based on the model that we have defined previously, we have to consider four sets of hyperparameters: the kernel hyperparameters $\theta$, the mean function parameters $\w$, the signal variance $\sigma_s^2$ and the noise variance $\sigma_n^2$.

A fully Bayesian approach does not admit a closed form solution for most kernel functions. In the Bayesian optimization literature, the standard and most effective approach is to rely on an empirical Bayes approach, where some parameters are approximated by a point value --e.g.: $\theta$-- or directly fixed a priori --e.g.: $\sigma_n^2$-- while some other parameters are actually computed in closed form --e.g.: $(\w,\sigma_s^2)$-- and 

On the other hand, we can assign conjugate hyperpriors to some parameters of the model, that is, $\w$ and $\sigma_s^2$ and still get a closed form posterior. For example, let assume that the prior can be decompose such as $p(\w,\sigma_s^2) = p(\w \mid \sigma^2_s) p(\sigma^2_s)$. For example, as a normal inverse-gamma: $p(\w,\sigma_s^2)  \sim \N \left(\w_0, \sigma_s^2 \W \right) \IG \left(\alpha, \beta \right)$, a normal scaled-inverse-chi-squared:  $p(\w,\sigma_s^2) \sim \N \left(\w_0, \sigma_s^2 \W \right) \chi^{-2} \left( \nu, \sigma^2_0 \right)$ or a Jeffreys prior: $p(\w,\sigma_s^2) \sim \mathbf{1} \cdot \sigma_s^{-2}$.

In fact, the first and second option are different parametrizations of the same model. The Jeffreys prior can be consider as the limit case. Also the Jeffreys prior \cite{Jeffreys1946}, is an \emph{uninformative} prior which is invariant to reparametrization --for example: we can exchange the signal variance $\sigma_s^2$ for the signal precision $\lambda_s$ and the prior is the same--. However, they can perform poorly for multidimensional parameters, for example, if they are many mean basis functions $\phi_i(\x)$.

\subsection{Learning the kernel parameters.}
For Bayesian optimization this can not be trivially implemented. As \cite{Vazquez10} proved, updating the value of $\theta$ during the optimization process introduce bias which might result of the optimization being stuck in a local minimum. To avoid this problem, we learn the parameters $\theta$ during a preliminar stage, with a small number of data points and then freeze or we updating the parameters infrequently.

We consider that the hyperprior of the kernel hyperparameters $\theta$ --if available-- is independent of other variables. Depending on the model, the likelihood function will be a multivariate Gaussian distribution or multivariate t distribution. Based on \cite{Santner03,Rasmussen:2006}, we are going to consider the following algorithms to learn the kernel hyperparameters:
\begin{itemize}
\item Cross-validation: In this case, we try to maximize the average predicted log probability by the \emph{leave one out} (LOO) strategy \cite{Rasmussen:2006}.
\item Maximum likelihood: For any of the models presented, one approach to learn the hyperparameters is to maximize the likelihood of all the parameters $\w$, $\sigma_s^2$ and $\theta$. Then, the likelihood function is a multivariate Gaussian distribution. As presented in \cite{Santner03}, we can obtain a better estimate if we adjust the number of degrees of freedom.
\item Posterior maximum likelihood: In this case, the likelihood function is modified to consider the posterior estimate of $(\w,\sigma_s^2)$ based on the different cases defined in Section \ref{sec:comp-hyperp}. In this case, the function will be a multivariate Gaussian or t distribution, depending on the kind of prior used for $\sigma_s^2$.
\item Maximum a posteriori: We can modify any of the previous algorithms by adding a prior distribution $p(\theta)$. As commented in \cite{Bull2011,Vazquez10}, applying the maximum likelihood naively while doing optimization might be problematic. \cite{ZiyuWang2013} suggest to add bounds to restrict the exploration of an overfitted or underfitted width parameter. However, this solution is restricted to using a box-bounded optimizer to learn $\theta$ and does not generalize to include extra information --if available-- about the parameters. Adding a prior distribution is a more general and elegant solution.
\item Sampling strategies: The applicability of those techniques seems to be against the philosophy of Bayesian optimization where we aim to reduce the number of data points. Nevertheless, some efforts have been developed in this area, as that presented in \cite{Snoek2012} or \cite{kleijnen2012expected}.
\end{itemize}

\section{BayesOpt Toolbox}

We present a highly efficient Bayesian optimization toolbox. It is implemented in C++ and it is compatible with Windows, Mac~Os and Linux. It also provides interfaces to C, Python and Matlab/Octave.

The library was influenced by the \emph{GPML} toolbox by \cite{Rasmussen2010} and NLOPT by \cite{Johnson}. The toolbox is highly configurable. The user can select among different kernels, mean functions, models and optimization criteria.

\subsection{Advantage of the library design}
\begin{figure}
  \centering
    \includegraphics[width=0.7\columnwidth]{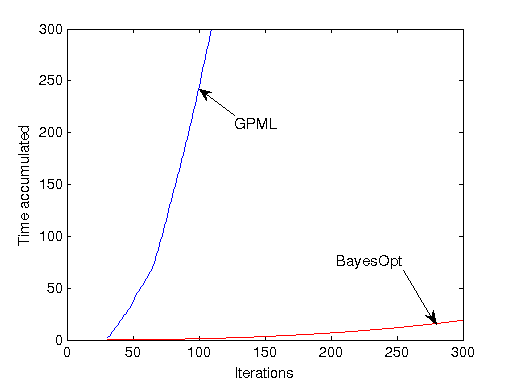}
  \caption{Time execution (in seconds) for the BayesOpt library compared to an equivalent Matlab naive implementation based on GPML. Clearly, the use of C/C++ for the core library combined with some code optimizations specially intended for iterative optimization reduce the total computational cost. The target function was a trivial function, thus the plot can be considered purely the cost of the optimization code.}
  \label{fig:time}
\end{figure}

First, it is \emph{fast}. The execution time of the library is represented in Figure \ref{fig:time} running on standard laptop and a single process. The optimization problem was trivial, so the computational cost plotted can be considered purely of the optimization process. 

One of the most critical components is the computation of the inverse of the kernel matrix. We have compared different numerical tricks like the incremental computation of the inverse matrix using blockwise inversion. We found that the \emph{Cholesky decomposition} method outperforms any other method in terms of performance and numerical stability. Besides, it guarantees the numerical stability of the Gram matrix.

Furthermore, we can use two properties of the method:
\begin{enumerate}
\item \emph{Points arrive one at a time.} Thus, we can do incremental computations of the matrix and vectors. For example, at each iteration, we know that only $n$ new elements will appear in the correlation matrix --the correlation of the new point with each of the existing points--. The rest of the matrix remains invariant. Thus, instead of computing the whole \emph{Cholesky} decomposition, being $\mathcal{O}(n^3)$ we just add the new row of elements to the triangular matrix, which is $\mathcal{O}(n^2)$. 
\item \emph{Multiple queries from the same model.} Computing the optimal decision requires to evaluate many queries. However, all those queries rely on the same previous data. Thus, we can factor out --precompute-- all the operations that are independent of the query.
\end{enumerate}

Second, it is \emph{flexible}. The design has been carefully selected relying on inheritance and polymorphism for all the components of the optimization process. It is very easy to create a new kernel, surrogate or criteria model by inheriting the abstract model or one of the existing models.  Thus, the newly created model will be fully integrated in the library. It will also inherit most of the existing linear algebra optimizations and safety checks. This is specially important in some part of the codes that have been carefully design to optimize performance or guarantee a correct implementation by design. For example, the initial set of points is selected using latin hypercube sampling. The kernel hyperparameters are not updated after every new data point since that is known to introduce bias \cite{Bull2011,Vazquez10}.

\begin{figure}
  \centering
  \includegraphics[width=.8\columnwidth]{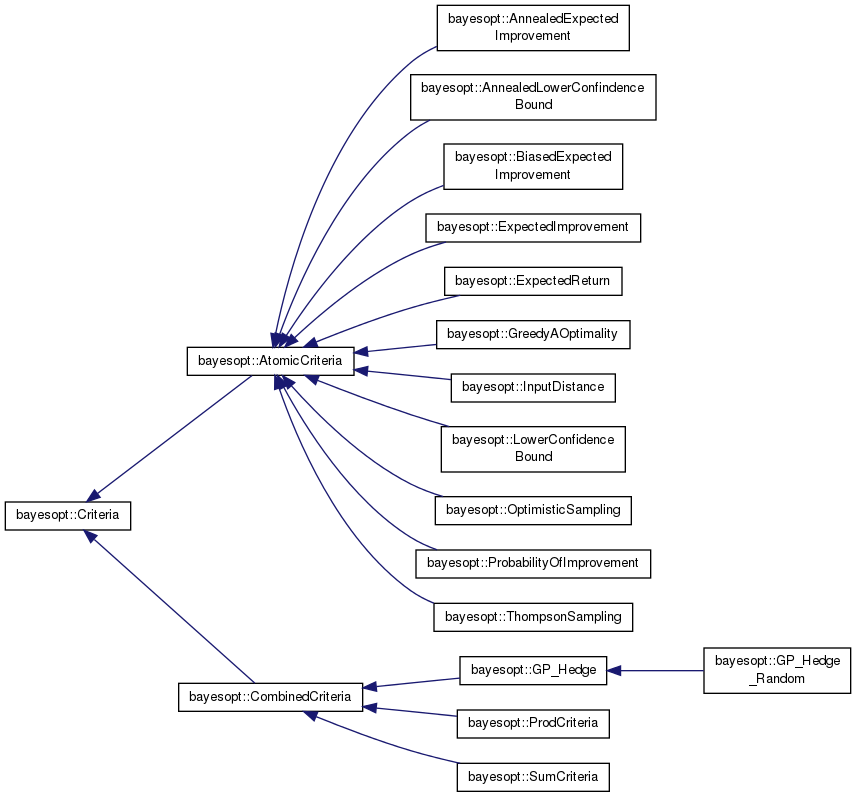}
  \caption{Inheritance diagram for the criteria class}
  \label{fig:inheritance}
\end{figure}

The library includes different kernel functions (Matern, Gaussian, isotropic, automatic relevance determination, etc.), different mean functions (constant, linear, radial, etc.), different surrogate models depending on the parameters hyperpriors (Gaussian process, Student's t process, etc.), and different criteria (expected improvement (EI), lower confidence bound (LCB), probability of improvement (POI), etc.). See Figure \ref{fig:inheritance} for the inheritance tree of the implemented criteria.

One of the advantages of having such a flexible design is that it is easy to combine different options. For example, we have implemented what we called, some metacriteria algorithms, like GP-Hedge \cite{Hoffman2011}, which can be used to find the most relevant criteria online. Other metacriterion is the linear combination of multiple criteria, which can be used to implement a optimization criteria with movement penalties \cite{Marchant2012}.

The library can be downloaded from
\url{https://bitbucket.org/rmcantin/bayesopt/downloads}
and the online documentation can be found in \url{http://rmcantin.bitbucket.org/html/}

The library internally uses NLOPT for the inner optimization loops (optimize criteria, learn kernel hyperparameters, etc.) \cite{Johnson}. The interface is very similar, thus NLOPT can be also used for comparison and benchmarking.

\subsection{Compatibility}
The toolbox has been design with the idea to be highly compatible in many platforms and setups. It has already been tested and compiled in different operating systems (Windows, Debian/Ubuntu, Mac~OS \ldots), with different compilers (Visual Studio, gcc, clang, mingw \ldots). The core of the library is in C++98 for compatibility with older compilers. It also provides interfaces and demos for C, Python and Matlab/Octave.

\subsection{Demos}
The code includes many demos for different languages, models (continuous, discrete), test functions (Ackley, Michalewicz, Rosenbrock, etc.). It even includes some specific demos such as multiprocessing computation or a simple computer vision application (image binarization).

\subsection{Using BayesOpt API}
The API is compatible with several languages and programming paradigms. In general, the usage of the toolbox can be summarize in three steps:
\begin{enumerate}
\item Define the function to optimize.
\item Set or modify the parameters and models of the optimization process.
\item Run the optimizer.
\end{enumerate}

Here we show a brief summary of the different ways to use the library:

\subsubsection{C/C++ callback usage}
This interface is the most standard approach, it could also be used as an interface or wrapper of other languages such as Fortran, Ada, etc.

The function to optimize must agree with the following template:
\begin{lstlisting}[language=C]
double my_function (unsigned int n, const double *x, 
                    double *gradient, void *func_data);
\end{lstlisting}
Then we call the optimizer for continuous or discrete optimization, passing the target function as a pointer\footnote{The gradient has been included for future compatibility. In the current implementation, it is not used.}.

\subsubsection{C++ inheritance usage}
This is the most straighforward and complete method to use the library. The object that must be optimized must inherit from the \texttt{bayesopt::ContinuousModel} or \texttt{bayesopt::DiscreteModel} classes. For example:
\begin{lstlisting}[language=C++]
class MyOptimization: public bayesopt::ContinuousModel
{
public:
 MyOptimization(bopt_params param): 
   ContinuousModel(param) {}

 double evaluateSample(const ublas::vector<double> &query) 
 {  // My function here   };

 bool checkReachability(const ublas::vector<double> &query)
 {  // My constraints here   };

 int optimize(ublas::vector<double> &result) 
};
\end{lstlisting}
Then, we just need to override one of the virtual functions called evaluateSample, which can be called with C arrays and uBlas vectors. Since there is no pure virtual functions, you can just redefine your preferred interface. You can also override the checkReachability function to include nonlinear constraints. Again, the parameters are defined in the \texttt{bopt\_params} struct.

\subsubsection{Python callback/inheritance usage}
Both interfaces are analogous to the C/C++ interface. In this case, the parameters are defined as a Python dictionary.

\subsubsection{Matlab/Octave callback usage}
Matlab/Octave only support callback. The parameters are defined as a Matlab struct analogous to the parameters struct in the C/C++ interface. 

\subsection{BayesOpt parameters}
The parameters are defined in the \texttt{bopt\_params} struct. The easiest way to set the parameters is to use
\begin{lstlisting}[language=C]
bopt_params parameters = initialize_parameters_to_default(); 
\end{lstlisting}
and then, modify the necesary fields. The details of each parameter can be found in the included documentation. For example:

\begin{lstlisting}[language=C++]
par.kernel.name = "kSum(kSEISO,kConst)"; // Sum of kernels
par.kernel.hp_mean = {0.5, 0.5};
par.kernel.hp_std = {1, 1};         // Hyperprior on kernel
par.kernel.n_hp = 2;
//Surrogate hyperpriors (w,sigma_s)
par.surr_name = S_STUDENT_T_PROCESS_JEFFREYS;
// We combine Expected Improvement, Lower Confidence Bound
// probability of improvement y Thompson sampling
par.crit_name = "cHedge(cEI,cLCB,cPOI,cThompsonSampling)";
par.crit_params = {1,1,0.01}; // Each criterion has 
par.n_crit_params = 3;        // its parameters
par.l_type = L_ML;         // Learning type
par.n_iterations = 200;    // Number of iterations
\end{lstlisting}

\begin{figure*}
\centering
\includegraphics[width=5cm]{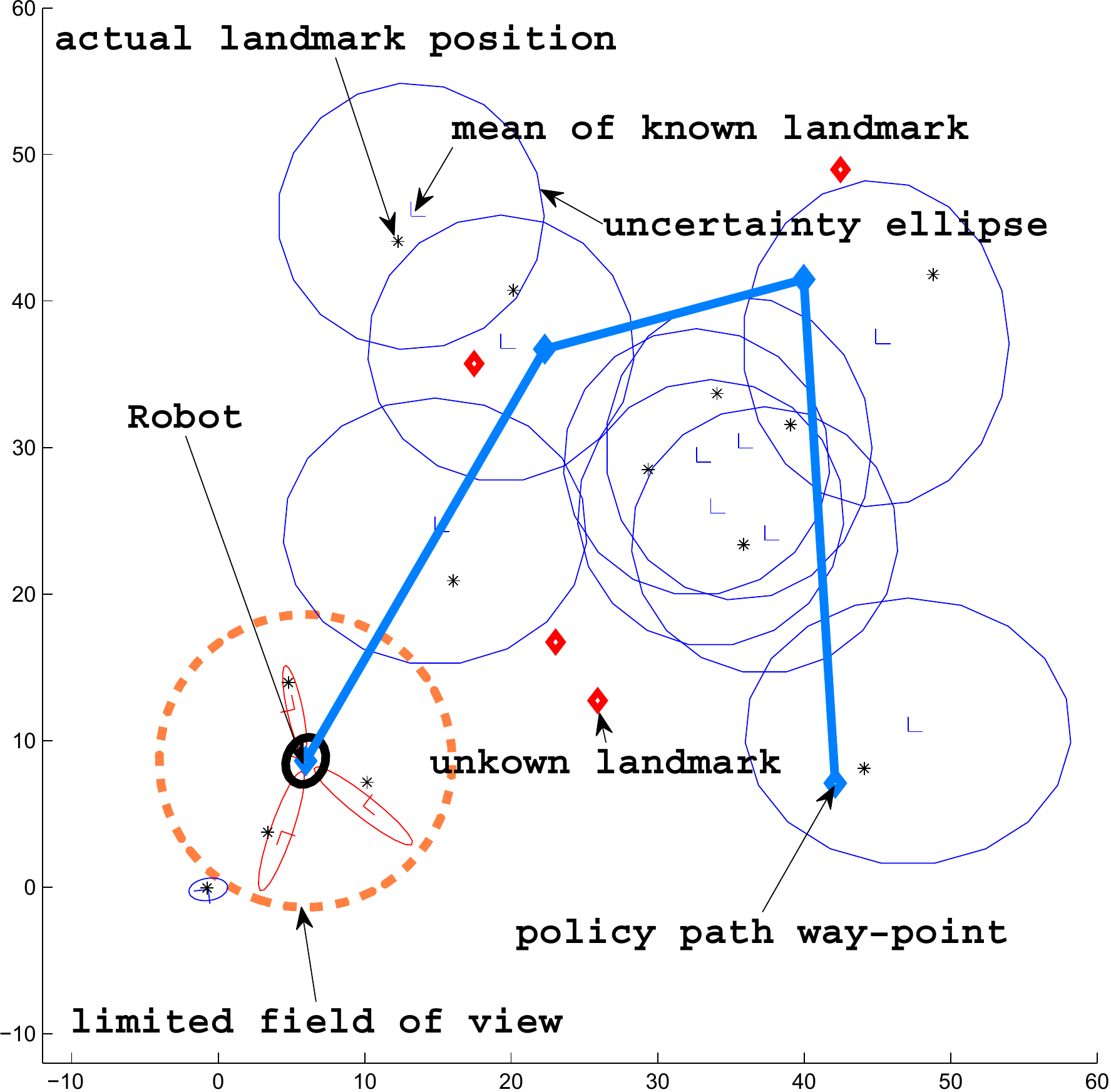}
\includegraphics[width=5cm]{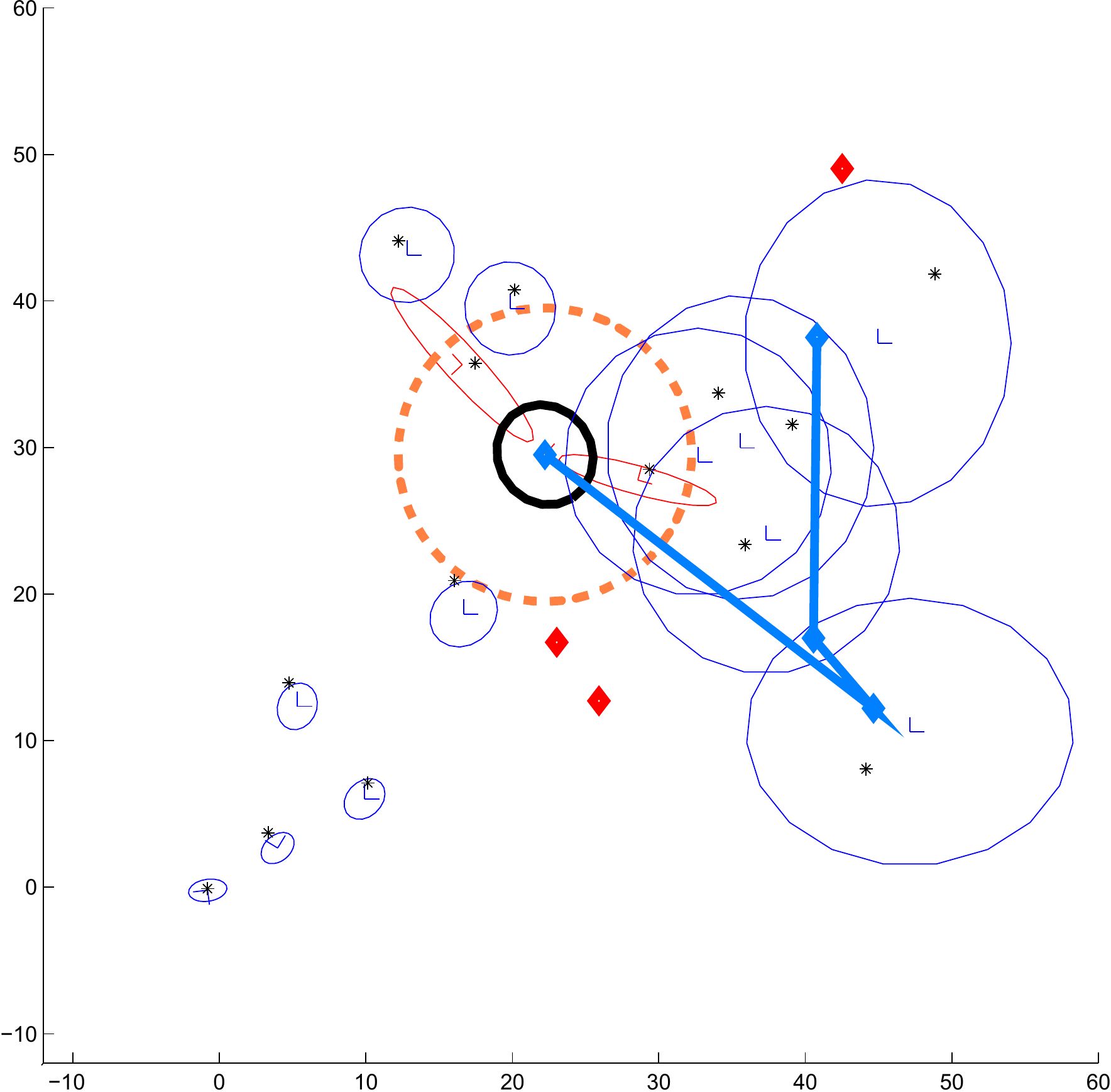}
\includegraphics[width=5cm]{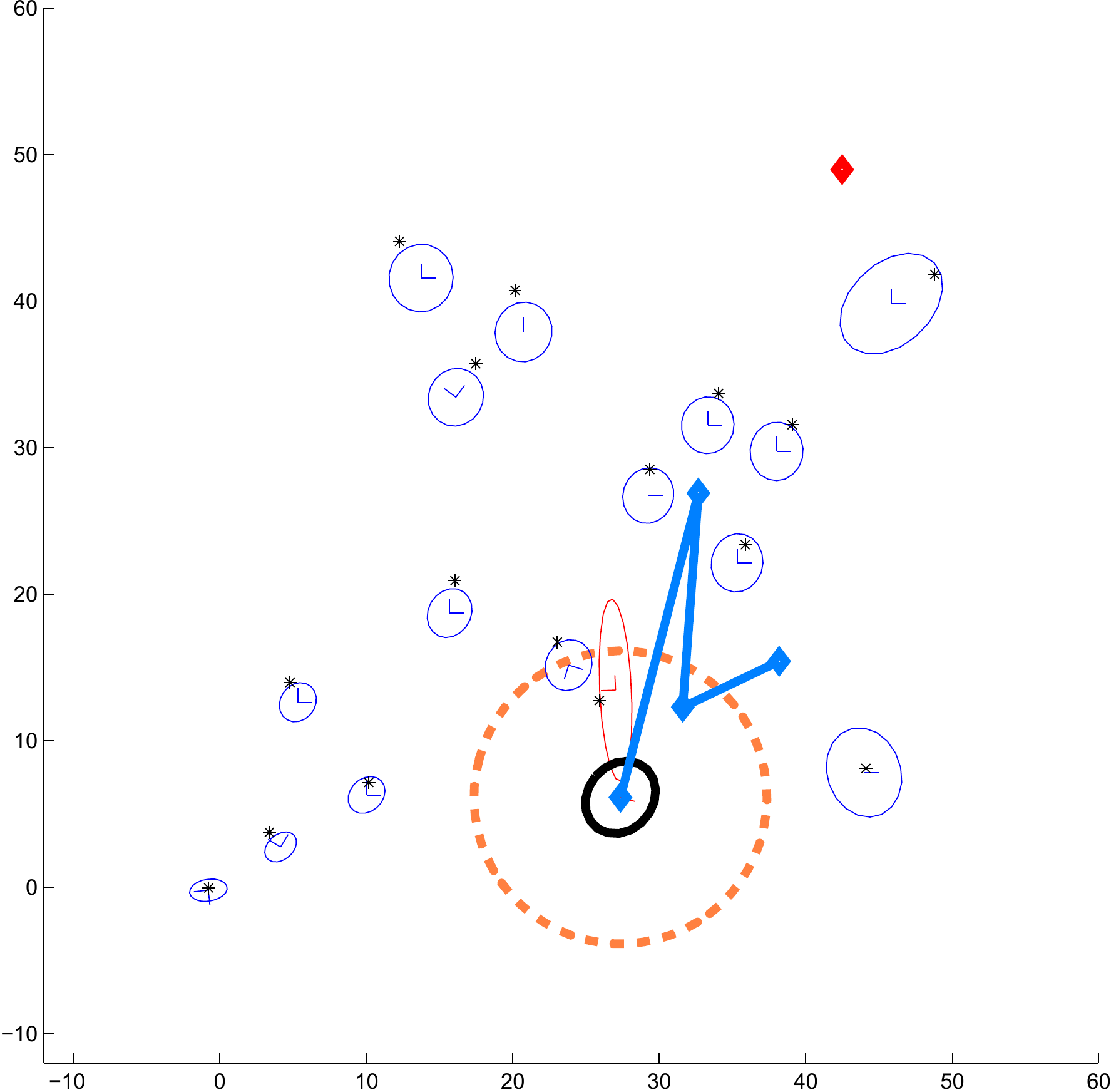}
\caption{This simulation shows three stages of the robot exploring an environment. The simulation includes landmarks that the robot does not know a priori. As soon as the robot observes these landmarks, it incorporates them into its model of the world. The robot continuously plans and replans so as to minimize the uncertainty in its pose and in the location of the known landmarks. The figure also shows the robot's limited field of view and the paths that it plans to follow at the three simulation stages.}
 \label{fig:introexploration}
\end{figure*}

\section{Robotics applications}
\label{sec:robot-appl}

Bayesian optimization has already been applied to mane robotics and related problems. It was first introduced in the field by \cite{Lizotte2007} and, independently by \cite{MartinezCantin07RSS}. The work of Lizotte et al. \cite{Lizotte2007} used Bayesian optimization for robot gait optimization, whereas Martinez-Cantin et al. \cite{MartinezCantin07RSS,MartinezCantin09AR} introduced a policy search algorithm for robot planning that assumes expensive reward functions (see Figure \ref{fig:introexploration}). Based on related work from \cite{Santner03}, the surrogate model in \cite{MartinezCantin07RSS} included hyperpriors on the model parameters. This work was the original motivation to build the library.

The idea of learning a controller and a planner was later combined and extended in \cite{Brochu:2010c} in a hierarchical fashion. Although it was originally intended for video games and simulation, its applicability to robotic platforms is direct.

As commented before, this topic is highly related with the work on sequential experimental design relying on Gaussian processes. In that field, one of the major breakthroughs was the discovery of the mutual information as an efficient criterion for submodular optimization. Originally intended for sensor placement \cite{Krause08JMLR} it was later extended for robot planning (moving sensors) \cite{singh09efficient,singh09nonmyopic}.

In fact, sensor placement or selection was addressed in a pure Bayesian optimization methodology using a more classical RMS error measure by Garnett et al. \cite{Garnett2010}.

A related problem of robots as \emph{moving sensors} was presented in \cite{Marchant2012}, which introduced the cost of robot movement as an additional criteria to trade-off exploration and exploitation.

Gait optimization in a more complex setup (snake robot) was further analyzed in \cite{Tesch_2011_7370} where the expected improvement algorithm is extended with non-controllable variables as in \cite{Williams_Santner_Notz_2000}. Recently, the same authors have extended the work to deal with multiobjective functions \cite{Tesch_2012_7375}.

Manipulation has also been attracted by the capabilities of Bayesian optimization. Kroemer et al. \cite{KroemerJRAS_66360} presented a reinforcement learning algorithm for grasping where the Bayesian optimization algorithm is used as a proxy between the reward function of the grasp and the grasping parameters. The reward function is directly computed by experimenting the grasp and performing a set of movements to guarantee stability of the grip. For the inner loop in the Bayesian optimization, the authors used a gradient-based strategy. This kind of strategy has some advantages in reinforcement learning for robots \cite{Peters_PIICIRS_2006}. However, they provide suboptimal performance due to local minima.

Instead of relying on arbitrary movements to test the grasp \emph{reward}, some authors have suggested a metric based on a simulated environment where the wrench space of the contact points can be computed \cite{Baier2007}. This idea was explored in the Bayesian optimization setup in \cite{Martinez-Cantin2012}. Later, it has been tested in \cite{Veiga2013} which uses the library presented here.

\begin{figure}
  \centering
  \includegraphics[width=\columnwidth]{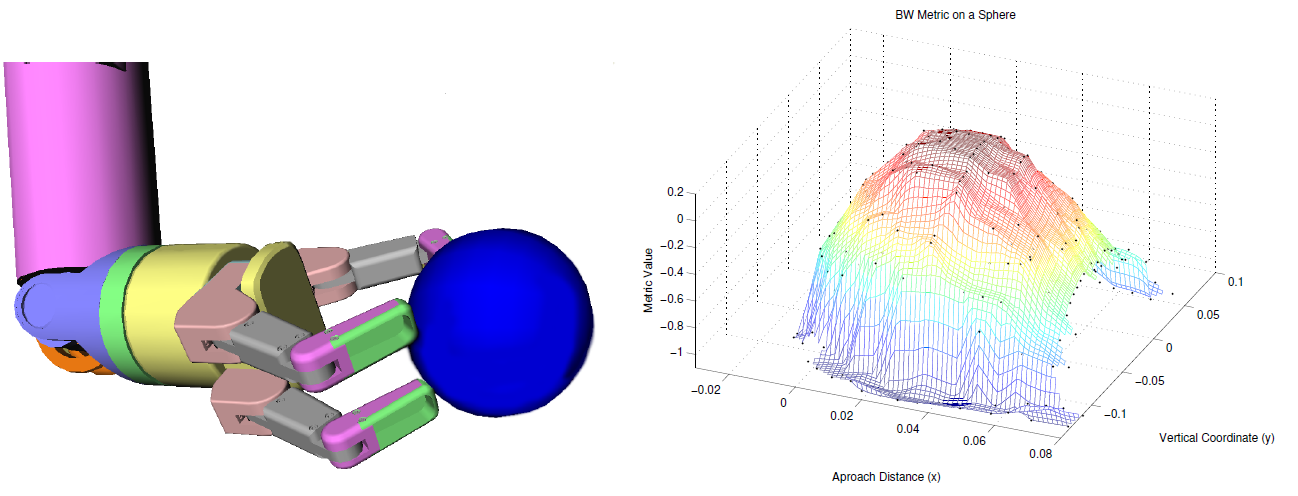}
  \caption{Bayesian optimization for grasping. The \emph{reward} of the grasp is computed by the volume of the wrench space generated by the contact points. This provides a metric of the stability of the grasp assuming that the friction of the fingers can be predicted.}
\end{figure}

Finally, as presented in a recent work \cite{HutHooLey11-smac}, the parameters  --either continuous, discrete or categorical-- of any algorithm can be optimized using Bayesian optimization.

\section{Conclusion} 
\label{sec:conclusion}

We have presented a common framework for Bayesian nonlinear optimization, sequential experimental design, bandits and, in general, hyperparameter optimization for many applications. The framework has been implemented in an efficient and easy to use library compatible with many operating systems and computer languages. It includes most of the state-of-the-art contributions to the field, both in terms of core algorithms and application oriented modifications.

As the implementation includes metamodels and metacriteria, the library is able to generalize to many situations with a small contribution from the user. On the other hand, for advanced users, it allows fully customization to improve the performance in dedicated applications.



\bibliographystyle{plainnat}
\bibliography{../bib/optimization}

\end{document}